\pdfoutput=1

\documentclass[11pt]{article}

\usepackage{emnlp2021}

\usepackage{times}
\usepackage{latexsym}

\usepackage[T1]{fontenc}

\usepackage[utf8]{inputenc}

\usepackage{microtype}

\usepackage{booktabs}
\usepackage[textsize=scriptsize]{todonotes}

\usepackage{amssymb}
\usepackage{amsmath}
\usepackage{graphicx}
\usepackage{enumitem}
\usepackage[dvipsnames]{xcolor}

\usepackage{varwidth}
\usepackage{tabularx}
\usepackage[marginal]{footmisc}

\usepackage{tikz}
\usetikzlibrary{positioning, arrows.meta, fit, calc}
\usepackage{pgfplots}
\pgfplotsset{compat=1.3}
\usepgfplotslibrary{fillbetween}

\title{Flexible Generation of Natural Language Deductions}

\author{Kaj Bostrom\ \ \ \ \ \ \ Xinyu Zhao\ \ \ \ \ \ \ Swarat Chaudhuri\ \ \ \ \ \ \ Greg Durrett\\
Department of Computer Science\\
  The University of Texas at Austin \\
  \texttt{kaj@cs.utexas.edu}}

\begin{document}
\maketitle
\begin{abstract}
An interpretable system for open-domain reasoning needs to express its reasoning process in a transparent form.
Natural language is an attractive representation for this purpose --- it is both highly expressive and easy for humans to understand.
However, manipulating natural language statements in logically consistent ways is hard: models must cope with variation in how meaning is expressed while remaining precise. In this paper, we describe \textsc{ParaPattern}, a method for building models to generate deductive inferences from diverse natural language inputs without direct human supervision. We train BART-based models \citep{lewis-etal-2020-bart} to generate the result of applying a particular logical operation to one or more premise statements. Crucially, we develop a largely automated pipeline for constructing suitable training examples from Wikipedia.
We evaluate our models using out-of-domain sentence compositions from the QASC \cite{qasc} and EntailmentBank \cite{entailmentbank} datasets as well as targeted perturbation sets.
Our results show that our models are substantially more accurate and flexible than baseline systems. \textsc{ParaPattern} achieves 85\% validity on examples of the `substitution' operation from EntailmentBank without the use of any in-domain training data, matching the performance of a model fine-tuned for EntailmentBank. 
The full source code for our method is publicly available.\footnote{\url{https://github.com/alephic/ParaPattern}}.
\end{abstract}


\section{Introduction}


Developing models that can make useful inferences from natural language premises has been a core goal in artificial intelligence since the field's early days \citep{bobrow_student, winograd_shrdlu}. Since then, there has been massive progress in automated formal reasoning~\citep{10.1145/1995376.1995394}; in contrast, progress in automated natural language reasoning has been comparatively slow. Today, `natural language inference' usually means recognizing textual entailment (RTE), a pairwise sentence classification task. Models have saturated RTE benchmarks \citep{bowman-etal-2015-large, williams-etal-2018-broad} largely through surface-level heuristics \cite{gururangan-etal-2018-annotation,poliak-etal-2018-hypothesis}; hill-climbing on these benchmarks has failed to yield robust models \cite{naik-etal-2018-stress} or systems capable of more complex reasoning.

\begin{figure}
    \centering
    \begin{tikzpicture}
    [
        font=\small,
        premise/.style={draw, rounded corners=2pt, fill=cyan, fill opacity=0.05, text opacity=1},
        model/.style={draw, rounded corners=2pt, fill=orange, fill opacity=0.05, text opacity=1},
        goal/.style={draw, rounded corners=2pt, fill=green, fill opacity=0.05, text opacity=1}
    ]
        
        \node[premise] at (-2, 0) (prem0) {\begin{varwidth}{3cm}The Green Bay Packers play in the NFL.\end{varwidth}};
        \node[premise] at (1.5, 0) (prem1) {\begin{varwidth}{3cm}NFL teams play from September to January.\end{varwidth}};
        \node[model] at (-1.5, -1) (model0) {\begin{varwidth}{3cm}\textit{Substitution model}\end{varwidth}};
        \node[goal] at (2, -1.25) (goal0) {\begin{varwidth}{3cm}The Green Bay Packers play from September to January.\end{varwidth}};
        
        \node[premise] at (-2, -2.25) (prem2) {\begin{varwidth}{3cm}Conditions lacking\\ environmental causes are strictly hereditary.\end{varwidth}};
        \node[model] at (-1.5, -3.5) (model1) {\begin{varwidth}{3cm}\textit{Contraposition model}\end{varwidth}};
        \node[goal] at (2, -3.25) (goal1) {\begin{varwidth}{3cm}Conditions that aren't strictly hereditary have environmental causes.\end{varwidth}};
        
        \draw[-Stealth] (prem0) -- (model0);
        \draw[-Stealth] (prem1) -- (model0);
        \draw[-Stealth] (model0) -- (goal0);
        \draw[-Stealth] (prem2) -- (model1);
        \draw[-Stealth] (model1) -- (goal1);

    \end{tikzpicture}
    \caption{Examples of the natural deduction operations for which we construct models. Note that conclusions involve both lexical inferences ($X$ \emph{plays in the NFL} $\rightarrow$ $X\ \text{is an}\ \textit{NFL team}$, $\lnot \lbrack X\ \textit{lacks}\ Y \rbrack\ \rightarrow\ X\ \textit{has}\ Y$) and logical transformations.}
    \label{fig:operation_examples}
\end{figure}
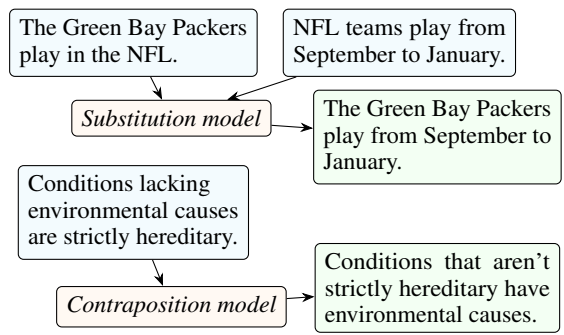

Following a line of work on multi-hop question answering \cite{welbl-etal-2018-constructing,yang-etal-2018-hotpotqa,chen-durrett-2019-understanding,min-etal-2019-compositional}, the reading comprehension community has started to make inroads in the area of reasoning. Recent datasets have been explicitly designed to test deduction ability \citep{liu2020natural, Yu2020ReClor, holzenberger2020dataset} and new types of models take inspiration from formal and informal reasoning \citep{ruletakers, saha-etal-2020-prover, cartuyvels-etal-2020-autoregressive, betz2021thinking}. Many recent modeling efforts share a common motif of using intermediate fact chains to support their final predictions, but a major shortcoming is that these chains are either retrieved heuristically or generated freely from autoregressive language models, meaning they are not necessarily sound.
To enforce soundness, we envision future reasoning systems factoring the deduction process into a set of common operations, analogous to proof rules. Modeling the reasoning process in this way would grant the ability to generalize systematically to any problem that could be decomposed in terms of available operations, among other desirable properties \citep{rudin2018stop}.

In this work, we describe a \emph{generative} model for single-step deductive reasoning, building towards models capable of generating the range of logical transformations needed for the full reasoning process. We use a BART-based sequence-to-sequence model \cite{lewis-etal-2020-bart} to represent the distribution of valid conclusion statements conditioned on one or more premise statements. To make sound inferences, the model must be fine-tuned on well-formed training data. We describe a pipeline for crafting this data based on syntactic retrieval from Wikipedia, rule-based example construction, and automatic paraphrasing to increase diversity. Our hypothesis is that the logical regularities in the constructed examples will teach models to generate correct deductions, while paraphrasing coupled with the inductive bias from pretraining will regularize models, allowing them to robustly tolerate natural lexical and syntactic variation in their inputs.

We demonstrate our method's effectiveness by using it to create models for two distinct logical operations, \textit{substitution} and \textit{contraposition}, examples of which are shown in Figure \ref{fig:operation_examples}. Through experiments on manually-constructed English perturbation sets, as well as on the English Question Answering via Sentence Composition (QASC) and EntailmentBank datasets \citep{qasc, entailmentbank}, we show that our proposed data generation method leads to accurate and robust operation models. While baseline methods tend to default to trivial input copying and fail to assign significant likelihood to valid novel conclusions, we show that our operation models reliably generate consistent inferences. Evaluating our substitution model on fact compositions from the QASC and EntailmentBank datasets, we find that our method produces valid conclusions at rates equivalent to models trained on in-domain, human-annotated data, indicating that our method is a viable substitute for expensive direct supervision.

\section{Methods}

We consider an operation $G$, like our substitution example (Fig.~\ref{fig:operation_examples}), to be analogous to a proof rule, allowing one or more premise statements to be combined and transformed to yield a valid conclusion statement.
A model for $G$ places a distribution $p_G(y \mid x_0, \ldots, x_n)$ over conclusions $y$ conditioned on premises $x_i$.

We would like models to satisfy three criteria:
\begin{description}
    \item[Consistency:] Predicted outputs should be valid deductions from the model's inputs.
    \item[Robustness:] Models should be robust to linguistic variation present in their inputs.
    \item[Supervision economy:] A minimal amount of manual effort should be needed to construct a model for a new operation.
\end{description}
    
We choose to parameterize $p_G$ by fine-tuning pretrained sequence-to-sequence language models \citep{lewis-etal-2020-bart, raffel-etal-t5}. Fine-tuning pretrained models allows the resulting operations to successfully handle a wider variety of inputs by leveraging general linguistic knowledge gained during pretraining.

The three desired model criteria we have identified lead to two data collection balancing acts:
\begin{itemize}
    \item Model consistency and robustness improve with increased data quantity, quality, and diversity, but collecting a large amount of diverse, high-quality data presents a challenge.
    \item Variation in the data and even noise will improve model robustness, but too much noise will cause the trained model to be inconsistent.
\end{itemize}

Directly annotating such data is possible, but requires significant manual labor, either in the form of expert annotation or careful prompting and filtering of crowd annotations. While annotated resources already exist for certain domains \citep{qasc, hwang2020cometatomic}, this is not the case for most types of reasoning. Scraping data from free text only works if examples of the desired operation appear in the wild, which is generally not the case for concise well-formed deduction steps.
\citet{betz2020critical} use templates to generate logically consistent text for training language models; however, there is little need for diversity or naturalism in their data as it is exclusively used during pretraining for the purposes of transfer learning. \citet{proofwriter} use template-based natural language proofs to fine-tune transformer language models for reasoning; we include a model trained on their data as one of our baseline systems.

\subsection{Data Collection}

Our proposed method, \textsc{ParaPattern}, combines scraping, template-based generation, and automatic paraphrasing in order to achieve sufficient data diversity and quality with very little manual effort. 

\begin{figure}[t]
\centering
\begin{tikzpicture}
    [
        font=\tiny,
        sentence/.style={draw, rounded corners=0.1cm, fill=CadetBlue, fill opacity=0.2, inner sep=0pt, minimum height=0.2cm, minimum width=0.6cm},
        premise/.style={draw, rounded corners=0.1cm, fill=CadetBlue, fill opacity=0.2, inner sep=0pt, minimum height=0.2cm, minimum width=0.4cm}
    ]
    
    \node at (0, 0) (wikilogo) {\includegraphics[width=1.0cm]{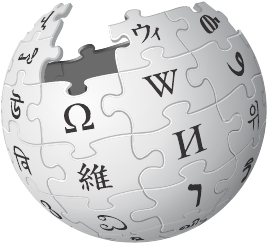}};
    \coordinate[right=0.0cm of wikilogo] (c0);
    \node[below=-0.2cm of wikilogo] {\begin{tabular}{@{}c@{}}\textbf{+}\\ pattern\end{tabular}};
    \coordinate[above=0.03cm of c0] (c1);
    \node[sentence, right=0.65cm of c1] (basesent0) {};
    \node[sentence, above=0.1cm of basesent0] (basesent1) {};
    \node[sentence, below=0.1cm of basesent0] (basesent2) {};
    \coordinate[left=0.1cm of basesent0] (c2);
    \draw[-Stealth] (c1) to node[midway, below=0.15cm] {\small $\mathbf{1}$} (c2);
    \coordinate (s1mid) at ($(basesent2)$);
    \node[below=0.2cm of s1mid] {\begin{tabular}{@{}c@{}}Base \\ sentences\end{tabular}};
    \node[premise, right=0.8cm of basesent0] (pa0) {};
    \draw[-Stealth, shorten <=0.1cm, shorten >=0.1cm] (basesent0) to node[midway, below=0.15cm] {\small $\mathbf{2}$} (pa0);
    \node[premise, above=0.1cm of pa0] (pa1) {};
    \node[premise, below=0.1cm of pa0] (pa2) {};
    \node[premise, right=0.1cm of pa0] (pb0) {};
    \node[premise, above=0.1cm of pb0] (pb1) {};
    \node[premise, below=0.1cm of pb0] (pb2) {};
    \node[below right=-0.05cm and -0.01cm of pa0, inner sep=0pt] {\small ,};
    \node[below right=-0.05cm and -0.01cm of pa1, inner sep=0pt] {\small ,};
    \node[below right=-0.05cm and -0.01cm of pa2, inner sep=0pt] {\small ,};
    \node[premise, right=0.2cm of pb0] (cn0) {};
    \node[premise, above=0.1cm of cn0] (cn1) {};
    \node[premise, below=0.1cm of cn0] (cn2) {};
    \draw[-{Classical TikZ Rightarrow}, shorten <=0.03cm, shorten >=0.03cm] (pb0) -- (cn0);
    \draw[-{Classical TikZ Rightarrow}, shorten <=0.03cm, shorten >=0.03cm] (pb1) -- (cn1);
    \draw[-{Classical TikZ Rightarrow}, shorten <=0.03cm, shorten >=0.03cm] (pb2) -- (cn2);
    \coordinate (s2mid) at ($(pa2)!0.5!(cn2)$);
    \node[below = 0.2cm of s2mid] {\begin{tabular}{@{}c@{}}Transformed\\ deduction examples\end{tabular}};
    
    \node[premise, right=0.8cm of cn0, fill=Magenta, fill opacity=0.4] (ppa0) {};
    \draw[-Stealth, shorten <=0.1cm, shorten >=0.1cm] (cn0) to node[midway, below=0.15cm] {\small $\mathbf{3}$} (ppa0);
    \node[premise, above=0.1cm of ppa0] (ppa1) {};
    \node[premise, below=0.1cm of ppa0, fill=NavyBlue, fill opacity=0.4] (ppa2) {};
    \node[premise, right=0.1cm of ppa0, fill=PineGreen, fill opacity=0.4] (ppb0) {};
    \node[premise, above=0.1cm of ppb0] (ppb1) {};
    \node[premise, below=0.1cm of ppb0, fill=Dandelion, fill opacity=0.4] (ppb2) {};
    \node[below right=-0.05cm and -0.01cm of ppa0, inner sep=0pt] {\small ,};
    \node[below right=-0.05cm and -0.01cm of ppa1, inner sep=0pt] {\small ,};
    \node[below right=-0.05cm and -0.01cm of ppa2, inner sep=0pt] {\small ,};
    \node[premise, right=0.2cm of ppb0] (pcn0) {};
    \node[premise, above=0.1cm of pcn0] (pcn1) {};
    \node[premise, below=0.1cm of pcn0] (pcn2) {};
    \draw[-{Classical TikZ Rightarrow}, shorten <=0.03cm, shorten >=0.03cm] (ppb0) -- (pcn0);
    \draw[-{Classical TikZ Rightarrow}, shorten <=0.03cm, shorten >=0.03cm] (ppb1) -- (pcn1);
    \draw[-{Classical TikZ Rightarrow}, shorten <=0.03cm, shorten >=0.03cm] (ppb2) -- (pcn2);
    \coordinate (s3mid) at ($(ppa2)!0.5!(pcn2)$);
    \node[below = 0.2cm of s3mid] {\begin{tabular}{@{}c@{}}Paraphrased\\ examples\end{tabular}};

\end{tikzpicture}
\caption{Schematic overview of the three phases of our data collection process: retrieval of base sentences from Wikipedia, expansion of these into reasoning examples, and paraphrasing.}
    \label{fig:schema}
\end{figure}

\begin{figure*}
    \centering
    \begin{tikzpicture}[
        font=\small,
        stage/.style={draw, rounded corners=2pt}
    ]
        \node[below] (t0) at (-1, -2.5) {\color{ForestGreen}\texttt{\textbf{NNS\$0}}};
        \node[below, right = 0.5cm of t0] (t1) {\texttt{\textbf{such}}};
        \node[below, right = 0.5cm of t1] (t2) {\texttt{\textbf{as}}};
        \node[below, right = 0.5cm of t2] (t3) {\color{NavyBlue}\texttt{\textbf{\$1}}};
        \node[below, right = 0.5cm of t3] (t4) {\color{Magenta}\texttt{\textbf{VBP\$2}}};
        \draw[bend left, -Stealth] (t0.north) to node [auto] (nsubj) {\tiny \texttt{nsubj}} (t4.north);
        \draw[bend right, -Stealth] (t2.north) to node [auto] {\tiny \texttt{prep}} (t0.north);
        \draw[bend right, -Stealth] (t3.north) to node [auto,swap] {\tiny \texttt{pobj}} (t2.north);
        \draw[bend right, -Stealth] (t1.south) to node [auto,swap] (amod) {\tiny \texttt{amod}} (t2.south);
        \node[above = 0.5cm of t4] (root) {\tiny \texttt{ROOT}};
        \draw (t4.north) -- (root.south);
        
        \node[stage, fit = (t0) (amod) (nsubj) (t4)] (patternbox) {};
        
        \node at (1.75, 0) (wikilogo) {\includegraphics[width=1.5cm]{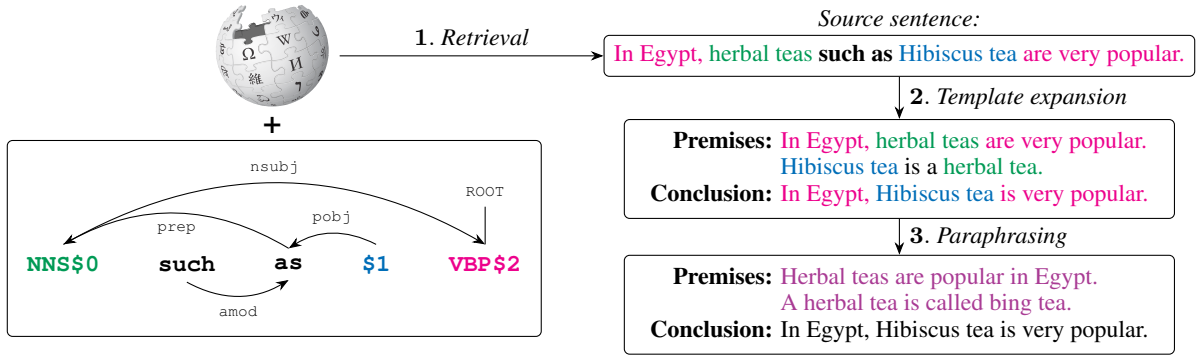}};
        
        \node[below = -0.15cm of wikilogo.south] {\large \textbf{+}};
        
        
        \node at (10, 0.5) {\textit{Source sentence:}};
        \node[stage] (s0) at (10, 0) {\begin{varwidth}{3in}{\color{Magenta} In Egypt,} {\color{ForestGreen} herbal teas} \textbf{such as} {\color{NavyBlue}Hibiscus tea} {\color{Magenta} are very popular.}\end{varwidth}};
        \node[stage] (s1) at (10, -1.5) {
        \begin{varwidth}{3in}
        \begin{tabular}{r@{ }l}
            \textbf{Premises:} & {\color{Magenta} In Egypt,} {\color{ForestGreen}herbal teas} {\color{Magenta}are very popular.}\\
            & {\color{NavyBlue}Hibiscus tea} is a {\color{ForestGreen}herbal tea.} \\
            \textbf{Conclusion:} & {\color{Magenta} In Egypt,} {\color{NavyBlue}Hibiscus tea} {\color{Magenta}is very popular.}\\
        \end{tabular}\end{varwidth}};
        \draw[-Stealth] (s0.south) to node [auto] {$\mathbf{2.}$ \textit{Template expansion}} (s1.north);
        \node[stage] (s2) at (10, -3.3) {
        \begin{varwidth}{3in}
        \begin{tabular}{r@{ }l}
            \textbf{Premises:} & {\color{Mulberry} Herbal teas are popular in Egypt.} \\
            & {\color{Mulberry}A herbal tea is called bing tea.} \\
            \textbf{Conclusion:} & In Egypt, Hibiscus tea is very popular.\\
        \end{tabular}\end{varwidth}};
        \draw[-Stealth] (s1.south) to node [auto] {$\mathbf{3.}$ \textit{Paraphrasing}} (s2.north);
        \draw[-Stealth] (wikilogo.east) to node [auto] {$\mathbf{1.}$ \textit{Retrieval}} (s0.west);
    \end{tikzpicture}
    
    \caption{An example of the steps involved in our data generation process for the substitution operation. Phrases in the source sentence and expanded template are colored according to their corresponding pattern variable.}
    \label{fig:template}
\end{figure*}

\textsc{ParaPattern} consists of three phases, as shown in Figures~\ref{fig:schema} and \ref{fig:template}.

\paragraph{Phase 1: Source scraping} A set of dependency patterns is used to retrieve source sentences suitable for template expansion from a dependency-parsed free text corpus. An example of one of the dependency patterns we use is shown in Figure~\ref{fig:template}. This template finds sentences exhibiting the Hearst pattern \cite{hearst-1992-automatic} \emph{X such as Y} indicating a hypernymy relationship between \emph{X} and \emph{Y}. Note that the retrieved sentences do not constitute complete training examples; such examples of logical reasoning are hard to find in the wild. These sentences need to be reshaped in the next step, but they are \emph{lexically diverse} and \emph{semantically suitable} as inputs to our templates in terms of the relations they express.
        
We perform syntactic search over a corpus of cleaned English Wikipedia article text comprising 112M 
sentences.
We use the off-the-shelf spaCy \texttt{en-core-web-sm} dependency parser \citep{spacy}, and index the resulting trees by bottom-up dependency chain prefixes in chunks of 160K sentences in order to accelerate the search process. Dependency parsing and index construction for English Wikipedia takes approximately 24 hours on a single CPU core.

We use six pattern variations to gather source sentences for the substitution template and two patterns for the contraposition template. Potential matches are filtered based on a list of disallowed subject modifiers that would result in semantically invalid examples. After filtering, the substitution patterns yield $\sim$44,000 source sentences and the contraposition patterns yield $\sim$23,000 source sentences. All dependency patterns we use are listed in Figure 7 in the appendix. Dependency search over the indexed trees takes 30-45 minutes depending on pattern complexity.
    
\paragraph{Phase 2: Template expansion} Source sentences are expanded into generated examples through the application of an operation-specific template. Figure \ref{fig:template} shows an example of a source sentence and its rule-based expansion into a pair of premise statements and a conclusion.
        
Template outputs are expressed in terms of the source pattern's match variables. The template expansion algorithm produces examples by breaking out dependency subtrees rooted at each match variable and rearranging them according to the template structure. We also apply simple heuristics for verb reinflection and noun number adjustment during the reconstruction process to maximize the fluency of the resulting text.

\begin{figure*}[tb!]
    \centering
    \scriptsize
    \begin{tabular}{@{}c c@{}}
    \begin{tabular}{@{}r@{ }l@{}}
    \multicolumn{2}{c}{\textbf{Substitution}} \\[1ex]
        \textbf{Premises:} & Staphylococcus epidermis is a microorganism.  \\
         & Microorganisms colonize the skin surface. \\
        \textbf{Paraphrased:} & {\color{Violet} Staphylococcus epidermidis is a microorganism.} \\
        & {\color{Violet} Microbiological colonization of the skin surface.} \\
        & {\color{VioletRed} ``Staphylococcus Epidermidis is a Microorganism.''} \\
        & {\color{VioletRed} The skin surface is colonized by micro organisms.} \\
        \textbf{Conclusion:} & Staphylococcus epidermis colonizes the skin surface. \\[1ex]
        \textbf{Premises:} & During the undergraduate years, seminarians learn the \\
        & ancient language courses. \\
        & Latin is an ancient language course. \\
        \textbf{Paraphrased:} & {\color{Violet} The seminars know the ancient language courses.} \\
        & {\color{Violet} Latin is an old language course.} \\
        & {\color{VioletRed} Seminarians learn ancient language during their} \\
        & {\color{VioletRed} undergraduate years.} \\
        & {\color{VioletRed} Latin is a language.} \\
        \textbf{Conclusion:} & During the undergraduate years, seminarians learn Latin.\\
    \end{tabular}
    &
    \begin{tabular}{@{}r@{ }l@{}}
    \multicolumn{2}{c}{\textbf{Contraposition}} \\[1ex]
        \textbf{Premise:} & As such, rivers that have headwaters in the mountains \\
        & provide water for irrigation in the surrounding lands. \\
        \textbf{Paraphrased:} & {\color{Mulberry} In order for water to be used in the surrounding lands,} \\
        & {\color{Mulberry}  the rivers in the mountains must have their headwaters there.} \\
        \textbf{Conclusion:} & As such, rivers that do not provide water for irrigation in the \\
        & surrounding lands do not have headwaters in the mountains. \\[1ex]
        \textbf{Premise:} & Dogs that are especially dirty or hungry are not able to \\
        & participate in contests. \\
        \textbf{Paraphrased:} & {\color{Mulberry} To participate in a contest, dogs that are dirty or hungry,} \\
        & {\color{Mulberry} must be turned away.} \\
        \textbf{Conclusion:} & Dogs that are able to participate in contests are not especially \\
        & dirty or hungry. \\
    \end{tabular}
    \end{tabular}
    \caption{Examples produced by our data generation pipeline.}
    \label{fig:data_exs}
\end{figure*}
    
\paragraph{Phase 3: Paraphrase augmentation} Data from template expansion is augmented by adding copies of each example with paraphrased input sentences. Paraphrases are generated using a version of the PEGASUS model \citep{zhang2019pegasus} fine-tuned for paraphrasing.\footnote{Model weights from \url{https://huggingface.co/tuner007/pegasus_paraphrase}} We sample two paraphrases for each original input using nucleus sampling with $p = 0.9$. See Figure \ref{fig:data_exs} for samples of input sentences after paraphrasing has been applied. These values were determined heuristically in the course of our preliminary experiments; we found that using a higher sampling cutoff or more paraphrases critically reduced the consistency of model inferences, and lowering $p$ or using only a single paraphrase per source example increased the rate of premise copying for examples not matching a training template.

We observe that the resulting paraphrases tend to include a noticeable amount of noise (e.g. the replacement of `Hibiscus' with `bing' in Figure \ref{fig:template}), but we hypothesize that since we only paraphrase premises, this effectively adds a denoising component to the fine-tuning objective similar to the motivation behind backtranslation in machine translation \cite{sennrich-etal-2016-improving}. Additional samples of the output of our data generation pipeline are shown in Figure \ref{fig:data_exs}. These examples demonstrate the ability of automatic paraphrasing to reduce both lexical and syntactic regularities in the original template outputs that can lead models to overfit to the template. In our subsequent experiments, we examine this overfitting by ablating Phase 3 of our data generation pipeline.

\subsection{Model Training} \label{sec:training}

Once data for an operation has been generated, we use it to fine-tune an instance of BART-Large \citep{lewis-etal-2020-bart}. Premise sentences $x_0 \ldots x_n$ are concatenated in a random order and provided as input to the model's encoder, and the conclusion sentence $y$ is used as the target sequence for the decoder.

We use model and training algorithm implementations from the \texttt{transformers} library \citep{wolf-etal-2020-transformers}.
We fine-tune models for a single epoch using the \textsc{AdamW} optimizer \citep{LoshchilovH19} with initial learning rate 3e-5 and triangular learning rate decay. In our preliminary experiments, we found that fine-tuning models for more than a single epoch always produced detrimental overfitting. Models are trained using a total batch size of 16 split across two NVidia Titan RTX GPUs; with this configuration, training takes no more than an hour of wall clock time per model.

\begin{figure*}[tb!]
    \centering
    \scriptsize
    \begin{tabular}{@{}c c@{}}
    \begin{tabular}{@{}r@{ }l@{}}
    \multicolumn{2}{c}{\textbf{Substitution - Control}} \\
        \textbf{Premises:} & RSA is a cryptographic system.  \\
         & Cryptographic systems let people exchange messages securely. \\
        \textbf{Conclusion:} & RSA lets people exchange messages securely. \\
        \textbf{Predicted:} & {\color{OliveGreen} RSA lets people exchange messages securely.} \\[1ex]
    \multicolumn{2}{c}{\textbf{Link NP mismatch}} \\
        \textbf{Premises:} & RSA is a cryptographic system.  \\
         & {\color{Emerald} Encryption protocols} let people exchange messages securely. \\
        \textbf{Conclusion:} & RSA lets people exchange messages securely. \\
        \textbf{Predicted:} & {\color{OliveGreen} RSA allows people to exchange messages securely.} \\[1ex]
    \multicolumn{2}{c}{\textbf{Identity VP mismatch}} \\
        \textbf{Premises:} & {\color{Emerald} Dominant} cryptographic systems {\color{Emerald} include} RSA.  \\
         & Cryptographic systems let people exchange messages securely. \\
        \textbf{Conclusion:} & RSA lets people exchange messages securely. \\
        \textbf{Predicted:} & {\color{OliveGreen} RSA allows people to exchange messages securely.}\\[1ex]
    \multicolumn{2}{c}{\textbf{NP + VP mismatch}} \\
        \textbf{Premises:} & {\color{Emerald} Dominant encryption protocols include} RSA.  \\
         & Cryptographic systems let people exchange messages securely. \\
        \textbf{Conclusion:} & RSA lets people exchange messages securely. \\
        \textbf{Predicted:} & {\color{OliveGreen} RSA allows people to exchange messages securely.}\\[1ex]
    \multicolumn{2}{c}{\textbf{Number agreement}} \\
        \textbf{Premises:} & RSA is a cryptographic system.  \\
         & Cryptographic systems {\color{Emerald} shield web traffic from surveillance} \\
         & {\color{Emerald} and let people communicate} securely.  \\
        \textbf{Conclusion:} & RSA {\color{Emerald} shields web traffic from surveillance and lets people} \\
        & {\color{Emerald} communicate} securely. \\
        \textbf{Predicted:} & {\color{OliveGreen} RSA shields web traffic from surveillance and} {\color{Orange}\textbf{let}} {\color{OliveGreen} people} \\
        & {\color{OliveGreen} communicate securely.}
    \end{tabular}
    &
    \begin{tabular}{@{}r@{ }l@{}}
    \multicolumn{2}{c}{\textbf{Contraposition - Control}} \\
        \textbf{Premise:} & Pesticides that contain DDT have harmful effects on birds. \\
        \textbf{Conclusion:} & Pesticides that do not have harmful effects on birds do not \\
        & contain DDT. \\
        \textbf{Predicted:} & {\color{OliveGreen} Pesticides that do not have harmful effects on birds do not} \\
        & {\color{OliveGreen} contain DDT.} \\[1ex]
    \multicolumn{2}{c}{\textbf{Postnominal modifier mismatch}} \\
        \textbf{Premise:} & Pesticides {\color{Emerald} containing} DDT have harmful effects on birds. \\
        \textbf{Conclusion:} & Pesticides that do not have harmful effects on birds do not \\
        & contain DDT. \\
        \textbf{Predicted:} & {\color{OliveGreen} Pesticides that do not have harmful effects on birds do not} \\
        & {\color{OliveGreen} contain DDT.} \\[1ex]
    \multicolumn{2}{c}{\textbf{Prenominal modifier mismatch}} \\
        \textbf{Premise:} & {\color{Emerald} DDT-containing} pesticides have harmful effects on birds. \\
        \textbf{Conclusion:} & Pesticides that do not have harmful effects on birds do not \\
        & contain DDT. \\
        \textbf{Predicted:} & {\color{OliveGreen} Pesticides that do not have harmful effects on birds do not} \\
        & {\color{OliveGreen} contain DDT.} \\[1ex]
    \multicolumn{2}{c}{\textbf{Premise negation}} \\
        \textbf{Premise:} & Pesticides that contain DDT {\color{Emerald} aren't safe for} birds. \\
        \textbf{Conclusion:} & Pesticides that are safe for birds do not contain DDT. \\
        \textbf{Predicted:} & {\color{OliveGreen} Pesticides that are safe for birds do not contain DDT.}
    \end{tabular}
    \end{tabular}
    \caption{Aligned perturbation set examples for substitution (left) and contraposition (right), with corresponding predicted \textsc{ParaPattern} BART output samples. Perturbed portions of each example are shown in {\color{Emerald}turquoise}. Grammatical errors are shown in {\color{Orange}orange}.}
    \label{fig:perturbation_exs}
\end{figure*}

\section{Experiments}

\subsection{Baselines}

We compare models trained using our proposed method against three baselines.

Our first baseline system is an unmodified instance of the pretrained autoregressive GPT2-Large language model \citep{radford2019language}, prompted with operation premises followed by the elicitation prefix ``Therefore,'' (\textbf{Zero-shot GPT2}). This baseline, inspired by the zero-shot premise elaboration employed by \citet{betz2021thinking}, is intended to assess the likelihood of making consistent deductions under a general model of language with no logical specialization.

Our second baseline model is an instance of BART-Large fine-tuned to generate hypotheses from the MNLI dataset \citep{williams-etal-2018-broad} conditioned on their respective premises (\textbf{MNLI BART}). We train on all instances for which the gold label indicates entailment ($\approx$103K examples) with the same training configuration as our other models, detailed in \ref{sec:training}. We hypothesize that while this model may assign higher likelihood to valid conclusions than a general language model would, it will place much more probability mass on re-emitting premise statements due to the fact that high word overlap tends to be a common feature of RTE examples labeled as `entailment' \citep{zhou-bansal-2020-towards}.

Our third baseline model is an instance of BART-Large fine-tuned to generate proof steps from the ProofWriter dataset \citep{proofwriter} (\textbf{ProofWriter BART}). While this dataset contains a large number of English proof steps ($\approx$135K), the language used in its proofs is automatically generated from a limited template library and is thus highly constrained. We hypothesize that this model will be unable to generalize as a result.

On the QASC and EntailmentBank datasets, we additionally compare to BART-Large models fine-tuned on inference steps from each dataset's respective training split (\textbf{QASC BART} and \textbf{Ent.~Bank BART}). The QASC training set contains $\sim$8K crowd-annotated fact compositions, while the EntailmentBank training set contains $\sim$3K expert-annotated premise-conclusion steps.

\subsection{Perturbation sets}

First, in order to evaluate the accuracy of our models on the operations they were designed for, and to understand the degree to which they generalize when input statements deviate from their training patterns, we manually construct controlled perturbation sets for each operation.

Our substitution perturbation set consists of 75 examples evenly split across a control condition and four test conditions, and our contraposition perturbation set consists of 60 examples evenly split across a control condition and three test conditions (15 examples per condition).

Each example in a given test condition is constructed by perturbing a corresponding control example. This aligned structure allows us to evaluate the impact of a particular perturbation on model performance without the confounding effect of content variation that would be present if each condition were constructed independently. Samples of each perturbation condition are presented in Figure \ref{fig:perturbation_exs}.


\begin{table*}[htb!]
    \centering \small
    \begin{tabular}{c @{\hspace{1em}} c @{\hspace{1em}} r @{ } l @{\hspace{1em}} c @{\hspace{2em}} c @{\hspace{1em}} r @{ } l @{\hspace{1em}} c}
    \toprule
         & \multicolumn{4}{c @{\hspace{2em}}}{\textbf{Substitution}} & \multicolumn{4}{c}{\textbf{Contraposition}} \\
        \textbf{Model} & \begin{tabular}{@{}c@{}}\textbf{Ref.} \\ \textbf{PPL}\end{tabular} $\downarrow$ &  \multicolumn{2}{c}{\textbf{BLEURT $\uparrow$}} & \textbf{Valid\% $\uparrow$} & \begin{tabular}{@{}c@{}}\textbf{Ref.} \\ \textbf{PPL}\end{tabular} $\downarrow$ & \multicolumn{2}{c}{\textbf{BLEURT $\uparrow$}} & \textbf{Valid\% $\uparrow$} \\
    \midrule
        Zero-shot GPT2            & 3.52          &         -0.88 & $\pm$ 0.35          & \phantom{00}1           & 6.04          & -0.89         & $\pm$ 0.31          & \phantom{00}1 \\
        MNLI BART                 & 2.00          &         -0.06 & $\pm$ 0.13          & \phantom{00}6           & 4.50          & -0.16         & $\pm$ 0.04          & \phantom{00}2 \\
        ProofWriter BART          & 3.86e2        &         -1.18 & $\pm$ 0.14          & \phantom{00}4           & 5.83e3        & -1.39         & $\pm$ 0.12          & \phantom{00}0 \\
        Pattern-only BART         & 3.55          &          0.49 & $\pm$ 0.01          & \phantom{0}55           & 3.16          & 0.31          & $\pm$ 0.00          & \phantom{0}38 \\
        \textsc{ParaPattern} BART & \textbf{1.54} & \textbf{0.66} & $\pm$ \textbf{0.05} & \textbf{\phantom{0}87}  & \textbf{1.57} & \textbf{0.69} & $\pm$ \textbf{0.07} & \textbf{\phantom{0}80} \\
    \bottomrule
    \end{tabular}
    \caption{Results for each perturbation set, averaged across perturbation conditions. Ref.~PPL refers to the perplexity of the reference conclusion under the model distribution. BLEURT scores are averaged across 10 samples per example; $\pm$ indicates the standard deviation of the samples. Valid\% refers to the proportion of generated conclusions rated as valid and non-redundant following manual review. Separate results for each perturbation condition can be found in Table 4 in the appendix.}
    \label{tab:probe_results}
\end{table*}

\subsection{QASC and EntailmentBank}

Our perturbation sets are not necessarily ``in-domain'' for our models, but they still neatly fit the reasoning patterns we are targeting. To test our approach's applicability to data outside its training, we apply our substitution model to the fact compositions in the validation splits of the QASC and EntailmentBank datasets \citep{qasc, entailmentbank}.

QASC fact compositions were annotated by crowd workers as rationales for multiple-choice question answering problems. Since annotators combined facts with a certain question in mind, there is some amount of missing context for many QASC fact combinations.

The EntailmentBank dataset consists of a set of expert-annotated natural language proofs for elementary science question-answer pairs involving multi-step reasoning. Thanks to its trained annotators, EntailmentBank contains fewer spurious fact combinations than QASC.

\subsection{Evaluation Criteria}

We evaluate model performance on each dataset primarily through a manual assessment of conclusion validity. The first author placed generated conclusions into one of six categories:
\begin{description}
\item[Valid:] Conclusion is logically consistent with premises but does not trivially repeat them.
\item[Valid with minor grammar errors:] Conclusion is valid but includes minor syntactic errors such as bad verb inflection that do not hinder comprehension.
\item[Repeats premises:] Conclusion is a near-verbatim copy of one or more premise sentences.
\item[Unsupported:] Conclusion is technically true but does not logically follow from premises.
\item[Incompatible:] Conclusion contradicts premises or is inherently false.
\item[Incomprehensible:] Conclusion is difficult to interpret due to major ungrammaticality.
\end{description}
Model outputs were shuffled and annotated without knowledge of model identity to prevent rating bias. The last author reannotated a subset of QASC examples to validate the first author's annotations; there were minor differences in interpretation of the divisions between non-valid categories, but the relative proportion of conclusions rated as valid remained consistent between annotators.

We additionally compute the perplexity of reference conclusions under each model in order to assess the likelihood assigned to desired conclusions by each model's output distribution.

For the perturbation sets, we also report the BLEURT score \citep{sellam-etal-2020-bleurt} of generated conclusions with respect to reference conclusions.

\section{Results}

\begin{figure*}[htb!]
    \centering
    \begin{tikzpicture}
        \pgfplotsset{every tick label/.append style={font=\tiny}}
        \begin{axis}
        [
            extra y ticks = 0,
            extra y tick labels = ,
            extra y tick style = { grid style = solid },
            axis x line = bottom,
            axis y line = left,
            axis line style = {-},
            ylabel = {\tiny BLEURT},
            ylabel shift = -18pt,
            xlabel = {\tiny Rank},
            xlabel shift = -16pt,
            xtick = {1},
            width=1.5in,
            height=3in,
            ymin = -1,
            ymax = 1,
            ytick distance=0.25,
            enlarge x limits=false,
            xmin=1,
            xmax=926,
            extra x ticks = {926},
            extra x tick labels = { 926\phantom{926} }
        ]
            \addplot [name path=data, mark = none] table [col sep=tab, x=ID, y=score] {qasc_scores.tsv};
            
            \path [name path=axis] (axis cs:0,0) -- (axis cs:925,0);
            
            \addplot [
                fill=red,
                fill opacity=0.05
            ]
            fill between[
                of=data and axis,
                split,
                every segment no 0/.style={
                    fill=blue
                },
            ];
            
            \coordinate (one) at (axis cs:0,1) {};
            \coordinate (oneR) at (axis cs:1000,1) {};
            \coordinate (two) at (axis cs:27,0.7458) {};
            \coordinate (twoR) at (axis cs:3000,0.7458) {};
            \coordinate (three) at (axis cs:117,0.5006) {};
            \coordinate (threeR) at (axis cs:900,0.5006) {};
            \coordinate (four) at (axis cs:278,0.2502) {};
            \coordinate (fourR) at (axis cs:900,0.2502) {};
            \coordinate (five) at (axis cs:486,0.0023) {};
            \coordinate (fiveR) at (axis cs:1000,0.0023) {};
            \coordinate (six) at (axis cs:704,-0.3328) {};
            \coordinate (sixR) at (axis cs:880,-0.3328) {};
            \coordinate (seven) at (axis cs:845,-0.6609) {};
            \coordinate (sevenR) at (axis cs:3600,-0.6609) {};
            \coordinate (eight) at (axis cs:912,-0.9929) {};
            \coordinate (eightR) at (axis cs:1100,-0.9929) {};

        \end{axis}
        
        \tikzstyle{every node}=[font=\tiny, fill=white, draw, inner sep=2pt]
        
        \draw (one) circle (2pt) -- (oneR);
        \node[above=-0.1cm of oneR, align=left, right] {
            \begin{varwidth}{8in}
            \begin{tabular}{@{}l@{ }l@{}}
                \textbf{Premises:} & All cnidarians are aquatic. \\
                 & Cnidarians are also called coelenterates. \\
                \textbf{Reference:} & Coelenterates are aquatic. \\
                \textbf{Generated:} & \color{OliveGreen} Coelenterates are aquatic. \\
            \end{tabular}
            \end{varwidth}
        };
        
        \draw (two) circle (2pt) -- (twoR);
        \node[above=0.55cm of twoR, align=left, right] {
            \begin{varwidth}{8in}
            \begin{tabular}{@{}l@{ }l@{}}
                \textbf{Premises:} & A barometer is used to measure air pressure. \\
                 & Air pressure is a measure of the weight of the air. \\
                \textbf{Reference:} & A barometer measures the weight of the air. \\
                \textbf{Generated:} & \color{OliveGreen} A barometer measures the weight of the air. \\
            \end{tabular}
            \end{varwidth}
        };
        
        \draw (three) circle (2pt) -- (threeR);
        \node[above=0.2cm of threeR, align=left, right] {
            \begin{varwidth}{8in}
            \begin{tabular}{@{}l@{ }l@{}}
                \textbf{Premises:} & Bad weather decreases visibility while driving. Another dangerous weather phenomena that thunderstorms produce is hail. \\
                \textbf{Reference:} & Hail decreases visibility while driving. \\
                \textbf{Generated:} & \color{OliveGreen} Thunderstorms produce hail, which decreases visibility while driving. \\
            \end{tabular}
            \end{varwidth}
        };
        
        \draw (four) circle (2pt) -- (fourR);
        \node[above=0.0cm of fourR, align=left, right] {
            \begin{varwidth}{8in}
            \begin{tabular}{@{}l@{ }l@{}}
                \textbf{Premises:} & Volcanic ash in the atmosphere decreases the temperature by blocking the sun. Most volcanic ash consists of fine, dusty particles. \\
                \textbf{Reference:} & Fine, dusty particles erupted from volcanos decrease the temperature by blocking the sun. \\
                \textbf{Generated:} & \color{OliveGreen} Most volcanic ash consists of fine, dusty particles that block the sun, lowering the temperature in the atmosphere. \\
            \end{tabular}
            \end{varwidth}
        };
        
        \draw (five) circle (2pt) -- (fiveR);
        \node[below=0.2cm of fiveR, align=left, right] {
            \begin{varwidth}{8in}
            \begin{tabular}{@{}l@{ }l@{}}
                \textbf{Premises:} & Kidney failure may be treated with dialysis. Dialysis is a way of cleaning the blood with an artificial kidney. \\
                \textbf{Reference:} & Kidney failure may be treated by cleaning the blood with an artificial kidney. \\
                \textbf{Generated:} & \color{OliveGreen} Kidney failure often {\color{Orange}require} an artificial kidney. \\
            \end{tabular}
            \end{varwidth}
        };
        
        \draw (six) circle (2pt) -- (sixR);
        \node[above=-0.3cm of sixR, align=left, right] {
            \begin{varwidth}{8in}
            \begin{tabular}{@{}l@{ }l@{}}
                \textbf{Premises:} & A compass is used to navigate oceans. \\
                                   & And to the east is the Atlantic Ocean. \\
                \textbf{Reference:} & A compass is used to navigate the Atlantic. \\
                \textbf{Generated:} & \color{OliveGreen} The Atlantic Ocean to the east{\color{Orange},} is navigated using a compass. \\
            \end{tabular}
            \end{varwidth}
        };
        
        \draw (seven) circle (2pt) -- (sevenR);
        \node[above=0.4cm of sevenR, align=left, right] {
            \begin{varwidth}{8in}
            \begin{tabular}{@{}l@{ }l@{}}
                \textbf{Premises:} & Howling is used to warn other wolves by wolves. \\
                                   & Canis lupus is the same species that includes grey wolves. \\
                \textbf{Reference:} & Canis lupus howl. \\
                \textbf{Generated:} & \color{OliveGreen} Canis lupus uses howling as a warning to other wolves. \\
            \end{tabular}
            \end{varwidth}
        };
        
        \draw (eight) circle (2pt) -- (eightR);
        \node[above=0.15cm of eightR, align=left, right] {
            \begin{varwidth}{8in}
            \begin{tabular}{@{}l@{ }l@{}}
                \textbf{Premises:} & Mutualism is a symbiotic relationship in which both species benefit. \\
                                   & Males have individual territories Has mutually beneficial symbiotic relationship with oxpecker bird. [sic] \\
                \textbf{Reference:} & Oxpecker birds help other species benefit. \\
                \textbf{Generated:} & \color{OliveGreen} Males have mutualism with oxpecker {\color{Orange}bird}. \\
            \end{tabular}
            \end{varwidth}
        };
        
    \end{tikzpicture}
    \caption{BLEURT score profile of ParaPattern BART substitution outputs for fact combinations from the QASC development set. Sampled substitution model outputs and corresponding QASC annotations for a range of scores are shown to the right. Minor grammatical errors are indicated in {\color{Orange} orange}. Note that generated conclusions remain semantically coherent despite diverging from annotated references as BLEURT scores decrease.}
    \label{fig:qasc_results}
\end{figure*}
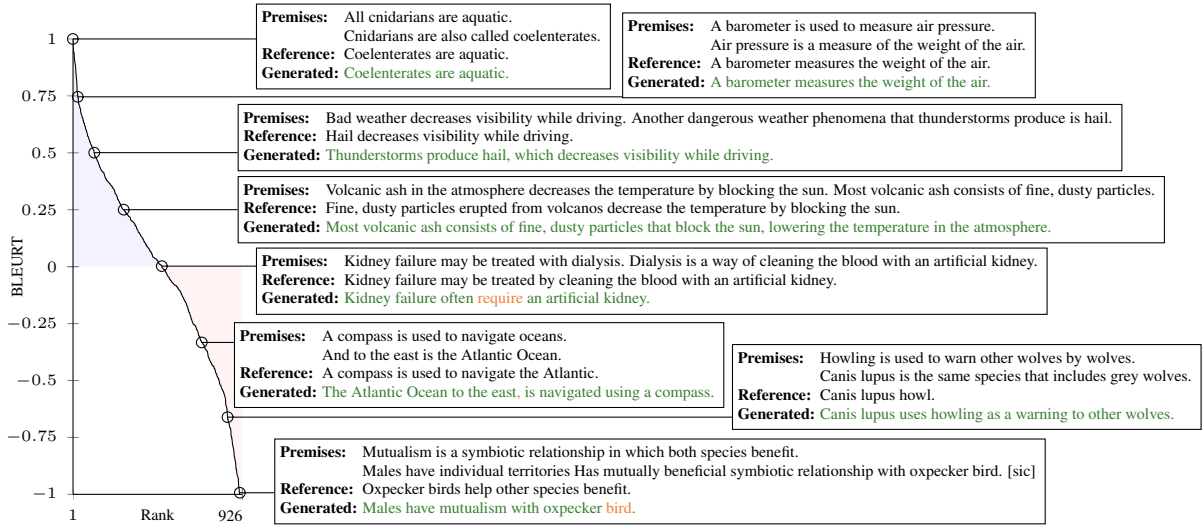

\subsection{Results on Perturbation Sets}

Our first question is whether or not we \textbf{have good generative models of natural language deductions.} As Table \ref{tab:probe_results} shows, \textsc{ParaPattern} BART outperforms all baselines by a wide margin in terms of the likelihood of desired conclusions (Ref.~PPL), the similarity of its outputs to desired conclusions, and its overall rate of valid inference. Additionally, there is a substantial gap in performance between models trained with and without paraphrastic data augmentation (\textsc{ParaPattern} vs.~Pattern-only, an ablation of our method). We observe that \textbf{\textsc{ParaPattern} allows models to reliably produce valid inferences} when given inputs that lie both on and off the training template manifold. In contrast, models trained using generic entailments (MNLI BART) or purely template-derived inferences (ProofWriter BART) are \emph{almost never} able to produce valid, non-redundant inferences.

\begin{table}[tb!]
\small
    \centering
    \begin{tabular}{c c c c}
        \toprule
        & \multicolumn{3}{@{}c@{}}{\textbf{QASC}}\\
        \textbf{Model} & \begin{tabular}{@{}c@{}}\textbf{Ref.}\\ \textbf{PPL}\end{tabular}$\downarrow$ & \textbf{Valid\%}$\uparrow$ & \textbf{Gram.\%}$\uparrow$\\
        \midrule
        Zero-shot GPT2 & 7.03 & \phantom{0}0 & \phantom{0}0 \\
        MNLI BART & \textbf{3.83} & \phantom{0}8 & \phantom{0}7 \\
        ProofWriter BART & 1.69e2 & \phantom{0}7 & \phantom{0}1 \\
        Ent. Bank BART & 6.61 & \textbf{72} & 62 \\
        Pattern-only BART & 39.7 & 16 & 10 \\
        \textsc{ParaPat.} BART & 4.82 & \textbf{73} & \textbf{68} \\
        \midrule
        QASC BART & 2.71 & 77 & 69 \\ \bottomrule
    \end{tabular}
    \caption{Results on the QASC development set. Valid\% indicates the proportion of predictions for 100 uniformly sampled examples that were rated as valid, non-redundant inferences following manual review. Gram.\% indicates the proportion of predictions rated both valid and free of grammatical errors.}
    \label{tab:qasc_numbers}
\end{table}

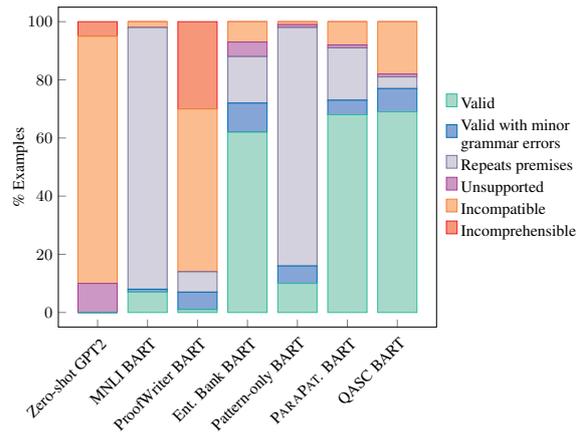
\begin{figure}[tb!]
\centering
\small
\resizebox{\columnwidth}{!}{
    \begin{tikzpicture}
    \begin{axis}[
        ybar stacked,
        bar width=22pt,
        x=28pt,
        enlarge y limits=0.05,
        enlarge x limits = {abs=22pt},
        legend style={at={(1.01, 0.5)}, anchor=west, font=\small, cells={align=left}, draw=none},
        legend cell align={left},
        ylabel={\% Examples},
        ylabel shift=-6pt,
        xtick={0,1,2,3,4,5,6},
        xticklabels={{Zero-shot GPT2}, {MNLI BART}, {ProofWriter BART}, {Ent. Bank BART}, {Pattern-only BART}, {\textsc{ParaPat.} BART}, {QASC BART}},
        x tick label style={rotate=45, anchor=east, yshift=-8pt, font=\small},
        tick pos = left,
        ymin=0,
        scale only axis = true,
        height=2.5in,
        width=3in
    ]
    \addplot+[ybar, SeaGreen, fill=SeaGreen!50!white] plot coordinates {(0,0) (1,7) (2,1) (3,62) (4,10) (5,68) (6,69)}; 
    \addplot+[ybar, NavyBlue, fill=NavyBlue!50!white] plot coordinates {(0,0) (1,1) (2,6) (3,10) (4,6) (5,5) (6,8)}; 
    \addplot+[ybar, CadetBlue, fill=CadetBlue!25!white] plot coordinates {(0,0) (1,90) (2,7) (3,16) (4,82) (5,18) (6,4)}; 
    \addplot+[ybar, Mulberry, fill=Mulberry!50!white] plot coordinates {(0,10) (1,0) (2,0) (3,5) (4,1) (5,1) (6,1)}; 
    \addplot+[ybar, Orange, fill=Orange!50!white] plot coordinates {(0,85) (1,2) (2,56) (3,7) (4,1) (5,8) (6,18)}; 
    \addplot+[ybar, Red, fill=Red!50!white] plot coordinates {(0,5) (1,0) (2,30) (3,0) (4,0) (5,0) (6,0)}; 
    
    \legend{\strut Valid, {\strut Valid with minor\\ grammar errors}, {\strut Repeats premises}, {\strut Unsupported}, {\strut Incompatible}, \strut Incomprehensible}
    
    \end{axis}
    \end{tikzpicture}
    }
    \caption{Detailed results of manual evaluation of QASC inferences for each model.}
    \label{fig:qasc_breakdown}
\end{figure}

\subsection{Results on QASC}

Table \ref{tab:qasc_numbers} shows that \textsc{ParaPattern} BART generates valid, grammatical inferences at a rate comparable to that of a model with identical parameter budget and pretraining fine-tuned on in-domain data (QASC BART) as well as a model fine-tuned on inferences from EntailmentBank (Ent.~Bank BART), an expert-annotated dataset in an adjacent domain. Thanks to our automated data collection pipeline, we are able to achieve this level of fidelity without any direct annotation.

For a full profile of model behaviors on QASC, refer to Figure \ref{fig:qasc_breakdown}.
We note that MNLI BART's preference for repeating premises results in a lower reference perplexity than \textsc{ParaPattern} BART in Table \ref{tab:qasc_numbers} despite MNLI BART behaving poorly during generation due to the fact that reference conclusions exhibit high lexical overlap with premises. Pattern-only BART also tends to repeat inputs, but in this case it is a low-confidence `fallback' behavior, as evidenced by its high reference perplexity.

\paragraph{Visualizing Model Generations on QASC} Figure \ref{fig:qasc_results} depicts a range of \textsc{ParaPattern} BART outputs for QASC validation set fact combinations ranked according to their BLEURT scores with respect to the reference combined fact. In the portion of this distribution above 0 BLEURT, we see very close agreement between the content of generated outputs and references. On the opposite end of the spectrum, we can see the structure of model outputs diverges from that of the reference fact combinations. However, even as our model's predictions grow farther from the reference conclusions, they remain semantically consistent combinations of the premise facts. The prediction for the final example in Figure \ref{fig:qasc_results} is a valid inference in spite of an ungrammatical premise, exemplifying one of the benefits of training on data augmented with noisily paraphrased inputs. In agreement with our quantitative results, these outputs confirm that \textbf{\textsc{ParaPattern} generates sound inferences even under domain shift}.

\begin{table}[tb!]
\small
    \centering
    \begin{tabular}{c c c c}
        \toprule
        & \multicolumn{3}{@{}c@{}}{\textbf{EntailmentBank}}\\
        \textbf{Model} & \begin{tabular}{@{}c@{}}\textbf{Ref.}\\ \textbf{PPL}\end{tabular}$\downarrow$ & \begin{tabular}{@{}c@{}}\textbf{Valid\%}\\ \textbf{(All)}\end{tabular}$\uparrow$ & \begin{tabular}{@{}c@{}}\textbf{Valid\%}\\ \textbf{(Subst.)}\end{tabular}$\uparrow$ \\
        \midrule
        \textsc{ParaPat.} BART & 4.70 & 57 & 85 \\
        Ent. Bank BART & 3.37 & 69 & 85 \\
        \bottomrule
    \end{tabular}
    \caption{Results on the EntailmentBank development set. Valid\% (All) indicates the proportion of predictions for 100 uniformly sampled examples manually rated as valid inferences. Valid\% (Subst.) indicates the proportion of valid predictions for the subset of examples classified as ``substitution'', as defined in \citet{entailmentbank}. This set includes 41\% of the sampled examples, in agreement with the statistic reported by the dataset's authors. We omit out-of-domain baselines due to their non-competitive performance on QASC.} 
    \label{tab:entailmentbank_numbers}
\end{table}

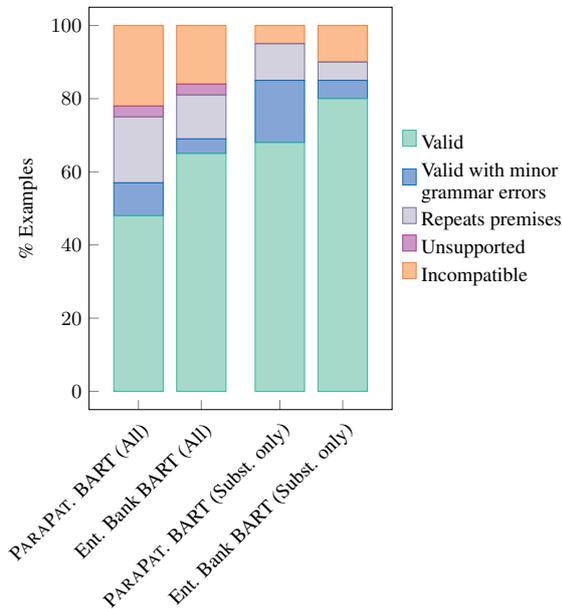
\begin{figure}[tb!]
    \centering
    \small
    \resizebox{\columnwidth}{!}{
    \begin{tikzpicture}
    \begin{axis}[
        ybar stacked,
        bar width=22pt,
        x=28pt,
        enlarge y limits = 0.05,
        enlarge x limits = {abs=22pt},
        legend style={at={(1.01, 0.5)}, anchor=west, font=\small, cells={align=left}, draw=none},
        legend cell align={left},
        ylabel={\% Examples},
        xtick={0,1,2.25,3.25},
        xticklabels={{\textsc{ParaPat.} BART (All)}, {Ent. Bank BART (All)}, {\textsc{ParaPat.} BART (Subst. only)}, {Ent. Bank BART (Subst. only)}},
        x tick label style={rotate=45, anchor=east, yshift=-8pt, font=\small},
        tick pos = left,
        ymin = 0,
        scale only axis = true,
        height=2.5in,
        width=2in
    ]
    \addplot+[ybar, SeaGreen, fill=SeaGreen!50!white] plot coordinates {(0,48) (1,65) (2.25,68) (3.25,80)}; 
    \addplot+[ybar, NavyBlue, fill=NavyBlue!50!white] plot coordinates {(0,9) (1,4) (2.25,17) (3.25,5)}; 
    \addplot+[ybar, CadetBlue, fill=CadetBlue!25!white] plot coordinates {(0,18) (1,12) (2.25,10) (3.25,5)}; 
    \addplot+[ybar, Mulberry, fill=Mulberry!50!white] plot coordinates {(0,3) (1,3) (2.25,0) (3.25,0)}; 
    \addplot+[ybar, Orange, fill=Orange!50!white] plot coordinates {(0,22) (1,16) (2.25,5) (3.25,10)}; 
    
    \legend{\strut Valid, {\strut Valid with minor\\ grammar errors}, {\strut Repeats premises}, {\strut Unsupported}, {\strut Incompatible}}
    
    \end{axis}
    \end{tikzpicture}
    }
    \caption{Detailed results of manual evaluation of EntailmentBank inferences for our proposed method (\textsc{ParaPat.} BART) and an in-domain fine-tuned model (Ent. Bank BART).}
    \label{fig:eb_breakdown}
\end{figure}

\subsection{Results on EntailmentBank}

We capitalize on the known taxonomy of reasoning types present in EntailmentBank to better understand how well our substitution model aligns with the definition of `substitution' adopted by the dataset's authors. In Table \ref{tab:entailmentbank_numbers} we demonstrate that \textsc{ParaPattern} BART matches the performance of the in-domain fine-tuned BART model on the subset of examples where the inputs specify a well-formed substitution as defined in \citet{entailmentbank}. This indicates that there is agreement between their definition of substitution, our model's representation of the operation, and the aspects of the operation captured by the EntailmentBank training set.

Furthermore, this result shows that \textbf{\textsc{ParaPattern} can circumvent the need for manual supervision for a given reasoning skill} without sacrificing performance in that skill.

Figure \ref{fig:eb_breakdown} provides a breakdown of model behaviors on EntailmentBank.
Of the 22\% of \textit{non-substitution} examples for which our substitution model is able to produce valid inferences, we note that the majority are instances of `property inheritance', `sequential inference', or `inference from rule', according to the taxonomy of \citet{entailmentbank}. The following are examples of some of these inferences generated by the \textsc{ParaPattern} BART substitution model:
\begin{description}
\setlength{\parskip}{0pt}
\item[Inference from rule] \hfill \\
\textit{If fossils are destroyed in rock transition, then there will be gaps in the fossil record. Fossils are lost / destroyed when sedimentary rock changes to metamorphic rock.}

$\rightarrow$ { \color{OliveGreen} \textit{Fossils are lost / destroyed when sedimentary rock changes to metamorphic rock, leaving gaps in the fossil record.}}
\item[Property inheritance] \hfill \\
\textit{United states is located in the northern hemisphere. New york / new york state is a state located in the united states of america.}

$\rightarrow$ { \color{OliveGreen} \textit{New york / new york state is in the northern hemisphere.}}
\item[Sequential inference] \hfill \\
\textit{Humans changing ecosystems usually has a negative impact on an ecosystem / organisms living in an ecosystem. Humans building homes in an ecosystem causes that ecosystem to change.}

$\rightarrow$ { \color{OliveGreen} \textit{Humans building homes in an ecosystem usually has a negative impact on an ecosystem / organisms living in an ecosystem.}}
\end{description}
We hypothesize that these inferences reflect generalizations of NP substitution to other phrase categories, most likely learned as a side effect of paraphrastic data augmentation.

\section{Related Work}

\textbf{Natural Logic} \cite{Bernardi2002,ZamanskyEtAl2006,MacCartneyManning2009,angeli-etal-2016-combining} is related to our approach in that it provides a framework for logical reasoning about statements in natural language. Such systems recognize that \emph{there is a cat on the dresser} entails \emph{there is an animal on the dresser} because of the hypernymy relationship between \emph{cat} and \emph{animal}. These relationships can be formalized into a monotonicity calculus \cite{IcardEtAl2017} and past work has grounded lexical inference tasks into such a formalism \cite{angeli-etal-2016-combining,hu-etal-2020-monalog}. Instead of decomposing entailment into relationships between words, our models learn to map premises to conclusions at the sentence level, allowing our approach to handle relationships not captured by such a formalism.

\paragraph{Multi-Hop Reasoning} Combining facts to form a conclusion overlaps with the idea of multi-hop reasoning, which has been explored in reading comprehension settings \cite{welbl-etal-2018-constructing,yang-etal-2018-hotpotqa}. However, training end-to-end models on these datasets does not necessarily teach models to combine facts \cite{chen-durrett-2019-understanding,min-etal-2019-compositional}. Systems like NLProlog \cite{weber-etal-2019-nlprolog} attempt to explicitly ground reasoning in logic, but this process still heavily relies on latent representations; in contrast, by grounding reasoning directly in natural language, a system based on natural deduction operations like ours gains inherent faithful natural language explanations and can build on the strengths of pretrained language models.

More recent datasets emphasize the ability to actually exhibit correct reasoning chains and form explanations \cite{ruletakers,xie-etal-2020-worldtree, entailmentbank}. Systems like PRover \cite{saha-etal-2020-prover} and Leap-of-Thought \cite{talmor-leap-of-thought} have some broadly similar goals to ours, but only \emph{retrieve} facts and do not generate novel conclusions.

\paragraph{Generative Reasoning} The generative nature of our models resembles generative models used for commonsense inference \cite{rajani-etal-2019-explain-custom,latcinnik2020explaining,shwartz-etal-2020-unsupervised}. However, these models do not strongly constrain the nature of what is generated. In contrast, our models reliably perform specific logical transformations, indicating that they can support sound inferences over longer reasoning chains in future work. \citet{arabshahi2021conversational} also explore generative reasoning in commonsense scenarios, but the domain of their approach is limited. \citet{khot2021text} use generative models to decompose a complex QA problem into a series of elementary steps that can be delegated to simpler models; this idea parallels the notion of decomposing reasoning into simple steps to be performed by generative operation models. Their results support the idea that such decomposition aids systematic generalization by enforcing separation of concerns.

\section{Conclusion}

Building systems that use natural language as a medium for reasoning will require operations to logically combine and transform natural language statements. In this work, we present \textsc{ParaPattern}, a method for creating such models with minimal manual effort by fine-tuning pretrained sequence-to-sequence language models on data generated through a three-step process of syntactic retrieval, template expansion, and automatic paraphrasing. Our experimental results show that \textsc{ParaPattern} yields operation models capable of generating consistent logical transformations over a diverse range of natural language inputs, matching the performance of models trained with in-domain human supervision.

\section*{Acknowledgments}

Thanks to Peter Clark for valuable discussion and to the anonymous reviewers for their helpful comments. This work was partially supported by NSF awards \# IIS-1814522 and \# CCF-1918651, by DARPA KAIROS award  \# FA8750-19-2-1003, and by ARO award \# W911NF-21-1-0009.


\bibliography{anthology,custom}
\bibliographystyle{acl_natbib}

\clearpage

\appendix

\onecolumn

\section{Appendix}
\label{sec:appendix_a}

\begin{figure}[h!]
\centering \small
\begin{tabular}{l}
    \textbf{Substitution source dependency patterns:}\\
     \texttt{[nsubj:NNS\$0 <[amod:`such' > prep:IN`as' < pobj:\$1]]> ROOT:VBP\$2}\\
     \texttt{[nsubj:NNS\$0 < prep:IN`like' < pobj:\$1]> ROOT:VBP\$2}\\
     \texttt{[nsubj:NNS\$0 < prep:VBG`include' < pobj:\$1]> ROOT:VBP\$2}\\
     \texttt{ROOT:VBP\$2 <[dobj:NNS\$0 <[amod:`such' > prep:IN`as' < pobj:\$1]]}\\
     \texttt{ROOT:VBP\$2 <[dobj:NNS\$0 < prep:IN`like' < pobj:\$1]}\\
     \texttt{ROOT:VBP\$2 <[dobj:NNS\$0 < prep:VBG`include' < pobj:\$1]}\\[1ex]
    \textbf{Contraposition source dependency patterns:}\\
    \texttt{[nsubj:NNS\$0 <[nsubj:WDT`that' > relcl:VBP\$1]] > ROOT:VBP\$2}\\
    \texttt{[nsubj:NNS\$0 <[prep:IN`with' < pobj:\$1]] > ROOT:VBP\$2}
\end{tabular}
\caption{All syntactic patterns used for data scraping. Pattern heads are written as \texttt{arclabel:POS`lemma'\$i}, where \texttt{arclabel} constrains the label on the arc to the matching token's head, \texttt{POS} constrains the part-of-speech tag of the matching token, and \texttt{`lemma'} constrains the lemmatized form of the matching token. \texttt{\$i} indicates that a matching token and the subtree under it will be exposed as a match variable for use in template expansion.}
\label{fig:scraping_patterns}
\end{figure}

\begin{table}[h!]
    \centering \small
    \begin{tabular}{c @{\hspace{1em}} c @{\hspace{1em}} r @{ } l @{\hspace{1em}} c @{\hspace{2em}} c @{\hspace{1em}} r @{ } l @{\hspace{1em}} c}
    \toprule
         & \multicolumn{4}{c @{\hspace{2em}}}{\textbf{Substitution}} & \multicolumn{4}{c}{\textbf{Contraposition}} \\
        \textbf{Model} & \begin{tabular}{@{}c@{}}\textbf{Ref.} \\ \textbf{PPL}\end{tabular} $\downarrow$ &  \multicolumn{2}{c}{\textbf{BLEURT $\uparrow$}} & \textbf{Valid\% $\uparrow$} & \begin{tabular}{@{}c@{}}\textbf{Ref.} \\ \textbf{PPL}\end{tabular} $\downarrow$ & \multicolumn{2}{c}{\textbf{BLEURT $\uparrow$}} & \textbf{Valid\% $\uparrow$} \\
    \midrule
     & \multicolumn{4}{c}{\textit{Control}} & \multicolumn{4}{c}{\textit{Control}} \\
        Zero-shot GPT2            & 3.28          &         -0.93 & $\pm$ 0.33          & \phantom{00}3           & 5.41          & -0.93         & $\pm$ 0.28          & \phantom{00}3 \\
        MNLI BART                 & 1.79          &          0.05 & $\pm$ 0.15          & \phantom{0}13           & 3.81          & -0.25         & $\pm$ 0.02          & \phantom{00}1 \\
        ProofWriter BART          & 1.72e2        &         -1.22 & $\pm$ 0.16          & \phantom{0}10           & 3.01e3        & -1.36         & $\pm$ 0.12          & \phantom{00}0 \\
        Pattern-only BART         & \textbf{1.01} & \textbf{0.89} & $\pm$ \textbf{0.00} & \textbf{100}            & \textbf{1.01} & \textbf{0.90} & $\pm$ \textbf{0.00} & \phantom{0}93 \\
        \textsc{ParaPattern} BART & 1.08          &          0.85 & $\pm$ 0.01          & \phantom{0}96           & 1.10          & \textbf{0.89} & $\pm$ \textbf{0.02} & \textbf{100} \\
    \midrule
     & \multicolumn{4}{c}{\textit{Link NP mismatch}} & \multicolumn{4}{c}{\textit{Postnominal modifier mismatch}} \\
        Zero-shot GPT2            & 3.61          &         -0.89 & $\pm$ 0.35          & \phantom{00}1           & 6.31          & -0.86         & $\pm$ 0.30          & \phantom{00}0 \\
        MNLI BART                 & 1.91          &         -0.04 & $\pm$ 0.07          & \phantom{00}0           & 4.79          & -0.29         & $\pm$ 0.02          & \phantom{00}8 \\
        ProofWriter BART          & 2.46e2        &         -1.21 & $\pm$ 0.20          & \phantom{00}9           & 3.25e3        & -1.42         & $\pm$ 0.14          & \phantom{00}0 \\
        Pattern-only BART         & 1.46          & \textbf{0.70} & $\pm$ \textbf{0.0}  & \phantom{0}53           & 2.23          & 0.00          & $\pm$ 0.00          & \phantom{00}0 \\
        \textsc{ParaPattern} BART & \textbf{1.39} & \textbf{0.68} & $\pm$ \textbf{0.05} & \textbf{\phantom{0}87}  & \textbf{1.39} & \textbf{0.75} & $\pm$ \textbf{0.08} & \textbf{\phantom{0}87} \\
    \midrule
     & \multicolumn{4}{c}{\textit{Identity VP mismatch}} & \multicolumn{4}{c}{\textit{Prenominal modifier mismatch}} \\
        Zero-shot GPT2            & 3.74          &         -0.87 & $\pm$ 0.36          & \phantom{00}2           & 6.96          & -0.87         & $\pm$ 0.28          & \phantom{00}0 \\
        MNLI BART                 & 2.17          &         -0.07 & $\pm$ 0.12          & \phantom{00}3           & 6.14          & -0.30         & $\pm$ 0.04          & \phantom{00}0 \\
        ProofWriter BART          & 2.68e2        &         -1.25 & $\pm$ 0.14          & \phantom{00}0           & 7.05e3        & -1.51         & $\pm$ 0.12          & \phantom{00}0 \\
        Pattern-only BART         & 4.39          &          0.09 & $\pm$ 0.00          & \phantom{0}13           & 7.08          & -0.37         & $\pm$ 0.00          & \phantom{00}0 \\
        \textsc{ParaPattern} BART & \textbf{1.59} & \textbf{0.52} & $\pm$ \textbf{0.14} & \textbf{\phantom{0}86}  & \textbf{1.79} & \textbf{0.48} & $\pm$ \textbf{0.15} & \textbf{\phantom{0}58} \\
    \midrule
         & \multicolumn{4}{c}{\textit{NP + VP mismatch}} & \multicolumn{4}{c}{\textit{Premise negation}} \\
        Zero-shot GPT2            & 4.15          &         -0.89 & $\pm$ 0.35          & \phantom{00}0           & 5.50          & -0.87         & $\pm$ 0.37          & \phantom{00}3 \\
        MNLI BART                 & 2.17          &         -0.18 & $\pm$ 0.17          & \phantom{00}2           & 3.24          & 0.20          & $\pm$ 0.07          & \phantom{00}0 \\
        ProofWriter BART          & 2.81e2        &         -1.31 & $\pm$ 0.08          & \phantom{00}0           & 1.00e4        & -1.26         & $\pm$ 0.11          & \phantom{00}0 \\
        Pattern-only BART         & 8.37          &          0.00 & $\pm$ 0.00          & \phantom{00}7           & 2.32          & \textbf{0.70} & $\pm$ \textbf{0.00} & \phantom{0}60 \\
        \textsc{ParaPattern} BART & \textbf{1.71} & \textbf{0.46} & $\pm$ \textbf{0.08} & \textbf{\phantom{0}75}  & \textbf{2.01} & 0.64          & $\pm$ 0.04          & \textbf{\phantom{0}75} \\
    \midrule
         & \multicolumn{4}{c}{\textit{Number agreement}} & & & &  \\
        Zero-shot GPT2            & 2.83          &         -0.81 & $\pm$ 0.33          & \phantom{00}1           & & & & \\
        MNLI BART                 & 1.97          &         -0.03 & $\pm$ 0.16          & \phantom{0}11           & & & & \\
        ProofWriter BART          & 9.63e2        &         -0.93 & $\pm$ 0.14          & \phantom{00}0           & & & & \\
        Pattern-only BART         & 2.53          & \textbf{0.77} & $\pm$ \textbf{0.00} & \textbf{100}            & & & & \\
        \textsc{ParaPattern} BART & \textbf{1.93} & \textbf{0.75} & $\pm$ \textbf{0.02} & \phantom{0}93           & & & & \\
    \bottomrule
    \end{tabular}
    \caption{Results for each perturbation set, broken down by test condition. Ref. PPL refers to the perplexity of the reference conclusion under the model distribution. BLEURT scores are averaged across 10 samples per example; $\pm$ indicates the standard deviation between samples. Valid\% refers to the proportion of generated conclusions rated as valid and non-redundant following manual review.}
    \label{tab:probe_results_full}
\end{table}

\end{document}